\documentclass[final]{siamltex}

\usepackage[]{fontenc}
\usepackage[latin1]{inputenc}
\usepackage{verbatim}
\usepackage{amsmath}
\usepackage{amssymb}
\usepackage{dsfont}
\usepackage{amsfonts}
\usepackage{mathrsfs}
\usepackage{graphicx}
\usepackage{color}
\usepackage{ulem}
\usepackage{xspace,bbm,epic,eepic}
\usepackage{pgfplots}
\usepackage{setspace}
\usepackage{xspace}

\usepackage{ stmaryrd }
\usepackage{graphicx}
\usepackage{lineno}
\usepackage{color}
\usepackage{url}
\usepackage{graphicx}
\usepackage{ stmaryrd }
\usepackage{algorithm}
\usepackage{algpseudocode}

\usepackage{amsmath,amssymb,color} 

\newcommand{\ie}{\textit{i.e.}, }

\DeclareMathOperator*{\argmin}{arg\,min}

\newtheorem{remark}{Remark}

\begin{document}

\title{Low-rank Approximation of Linear Maps}

\author{P. H\'EAS\thanks{INRIA Centre Rennes - Bretagne Atlantique, campus universitaire de Beaulieu, 35042 Rennes, France ({\tt patrick.heas@inria.fr, cedric.herzet@inria.fr}) }  \and   C. HERZET$^* $ 
   }

\date{}
\maketitle
\begin{abstract}
This work provides closed-form solutions and minimum achievable errors for a large class of low-rank approximation problems in Hilbert spaces. The proposed theorem generalizes to the case of bounded linear operators the previous results obtained in the finite dimensional case for the Frobenius norm. The theorem provides the basis for the design of tractable algorithms for kernel or continuous DMD.
\end{abstract}
\section{Introduction}
 Let  $\mathcal{U}$ and $\mathcal{V}$ be two separable Hilbert spaces of dimension $\textrm{dim}(\mathcal{U})$ and $\textrm{dim}(\mathcal{V})$, possibly infinite. Let    $\mathcal{B}(\mathcal{U},\mathcal{V})$ denote the class of linear bounded operators from  $\mathcal{U}$ to $\mathcal{V}$ and 
 $\mathcal{B}_k(\mathcal{U},\mathcal{V}) = \{  M \in \mathcal{B}(\mathcal{U},\mathcal{V}): \textrm{rank}(M)\le k\},$
where $\textrm{rank}(\cdot)$ denotes the rank  operator.
In this work, we are interested in characterizing the solutions  of the following constrained optimization problem
		\begin{align}\label{eq:prob0} 
	&\argmin_{M \in  \mathcal{B}_k(\mathcal{V},\mathcal{V}) } \|Y -M  \, X \|,\quad X,Y \in \mathcal{S}\subseteq \mathcal{B}(\mathcal{U},\mathcal{V}),
		\end{align} 
where $\|\cdot\|$ is some operator norm and  $\mathcal{S}$ some subset to be specified in due time.
Problem \eqref{eq:prob0} is non-convex due to the rank constraint and is in general infinite-dimensional. This raises the question of the tractability  of  problems of the form of \eqref{eq:prob0}. 

In the last decade, there has been a surge of interest for  low-rank solutions of linear matrix equations~\cite{fazel2002matrix,jain2010guaranteed,lee2009guaranteed,lee2010admira,mishra2013low,recht2010guaranteed}.  Problems of the form of \eqref{eq:prob0} can be viewed as generalizations to the infinite dimensional case of some of those matrix equations. In the finite dimensional case, certain instances with a  very special structures   admit closed-form solutions \cite{eckart1936approximation,HeasHerzet17,mesbahi1997rank,parrilo2000cone}, which can be computed in polynomial time. We mention that some authors have proposed tractable but sub-optimal solutions to some particular  finite~\cite{Chen12,Jovanovic12,Tu2014391,williams2015data} and infinite-dimensional \cite{williams2014kernel} problem instances.

In this work, we show that some infinite-dimensional problems of the form of \eqref{eq:prob0}  admit also a closed-form solution. The proof relies on the well-know  Schmidt-Eckhart-Young-Mirsky theorem \cite{schmidt1907theorie}. The theorem exposed in this work  can be viewed as a direct generalization to the infinite dimensional case of \cite[Theorem 4.1]{HeasHerzet17}. 

\section{Problem Statement and Solution}
We begin by introducing some notations,  then define the low-rank approximation problem and finally provide  a closed-form solution and error norm.
 
\subsection{Notations}
 Let  $\{e^\mathcal{U}_i\}_{i=1}^{\textrm{dim}(\mathcal{U})}$ and $\{e^\mathcal{V}_i\}_{i=1}^{\textrm{dim}(\mathcal{V})}$ be any ONBs of $\mathcal{U}$ and $\mathcal{V}$. The inner product  in those spaces will be denoted by $\langle \cdot ,\cdot \rangle_\mathcal{U}$ and $\langle \cdot ,\cdot \rangle_\mathcal{V}$ and their induced norm by $\| \cdot \|_\mathcal{U}$ and $\| \cdot \|_\mathcal{V}$ . Let $m=\min(\textrm{dim}(\mathcal{U}),\textrm{dim}(\mathcal{V})).$  Let $I \in \mathcal{B}(\mathcal{U}, \mathcal{U})$ be the identity operator.  Let the singular value decomposition (SVD) of $ M \in \mathcal{B}(\mathcal{U},\mathcal{V})$ be
$$M  =\sum_{i=1}^{m} \sigma_i^M  \varphi_i^M \langle \psi_i^M, \cdot \rangle_{\mathcal{U}} ,$$
where $\{\varphi^M_i\}_{i=1}^m$, $\{\psi^M_i\}_{i=1}^m$  are respectively the left and right singular functions associated to the  singular values  $\{\sigma^M_i\}_{i=1}^m$  of $M$ ~\cite{zhuoperator}. The pseudo inverse of $M$ denoted $ M^\dagger \in \mathcal{B}(\mathcal{V},\mathcal{U})$  will be  defined as
$$M^\dagger  =\sum_{i=1}^{m} (\sigma_i^M)^{\dagger}  \psi_i^M \langle \varphi_i^M, \cdot \rangle_{\mathcal{V}} ,\quad \textrm{where}\quad 
(\sigma_i^M)^{\dagger}=  \left\{\begin{aligned}
&(\sigma_i^M)^{-1}\quad \textrm{if}\quad  \sigma_i^M > 0\\
&0\quad\quad\quad \textrm{else}
\end{aligned}\right. \vspace{-0.cm}
.$$

 The the Hilbert-Schmidt norm, denoted by $\| \cdot \|_{\mathcal{HS}}$, is defined for any  $M \in \mathcal{B}(\mathcal{U},\mathcal{V})$ as
$$
\| M \|_{\mathcal{HS}} =\left( \sum_{i=1}^{\textrm{dim}(\mathcal{U})}\| M e^\mathcal{U}_i \|^2_\mathcal{V} \right)^{\frac{1}{2}}=\left( \sum_{i=1}^{m} (\sigma_i^M)^{2}   \right)^{\frac{1}{2}} ,
$$
and the class of   Hilbert-Schmidt operators is 
$\mathcal{S}(\mathcal{U},\mathcal{V}) = \{  M \in \mathcal{B}(\mathcal{U},\mathcal{V}): \|M\|_{\mathcal{HS}} < \infty\}$~\cite{zhuoperator}.

\subsection{Optimization Problem}
 We are now ready to clarify the definition of problem \eqref{eq:prob0}. Let $X,Y \in \mathcal{S}(\mathcal{U},\mathcal{V})$.  We are interested in the  low-rank approximation problem solutions 
		\begin{align}\label{eq:prob1} 
		M_k^\star \in &\argmin_{M \in  \mathcal{B}_k(\mathcal{V},\mathcal{V}) } \|Y -M \, X \|_{\mathcal{HS}}.
		\end{align}

\subsection{Closed-Form Solution}

We detail in the following theorem our  result. The proof is given in Section~\ref{ProofTH}.\\

\begin{theorem}\label{Theorem:1}
Problem \eqref{eq:prob1}  admits the optimal solution 
$$M_k^\star=P_k \, Y \,  X^\dagger,$$ where $P_k \in  \mathcal{B}_k(\mathcal{V},\mathcal{V})$ is given by
$ P_k  = \sum_{i=1}^k \varphi_i^Z \langle \varphi_i^Z  , \cdot \rangle_\mathcal{V}$ and $Z= Y \, X^\dagger  \, X $. Moreover, the square of the approximation error is
\begin{align}\label{eq:optError}
 \|Y -M_k^\star \, X \|_{\mathcal{HS}}^2=  \sum_{i=k+1}^{m} (\sigma_i^Z)^{2}  +  \sum_{i=\rank(X)+1}^m  \sum_{j=1}^m ( \sigma_j^Y )^2\langle \psi_j^Y, \psi_i^X \rangle_\mathcal{U}^2 ,
 \end{align}
where $m=\min(\textrm{dim}(\mathcal{U}),\textrm{dim}(\mathcal{V}))$.\\
 \end{theorem}

\begin{remark}[Modified  Hilbert-Schmidt  Norm]\label{remark:2}
The result can be extended for an approximation in the sense of the modified  Hilbert-Schmidt norm. In particular, for $m < \infty$, this extension can be seen as the DMD counterpart to the  POD  problem with energy inner product presented in \cite[Proposition 6.2]{quarteroni2015reduced}.
Let us define this modified norm.  We need first to introduce an additional norm for $\mathcal{V}$ induced by an alternative inner product. For any $v_1,v_2 \in \mathcal{V}$, we define 
$$ \langle v_1 ,v_2 \rangle_{\mathcal{V}_K} =  \langle v_1 , K \, v_2 \rangle_{\mathcal{V}},$$
where $K \in \mathcal{B}(\mathcal{V},\mathcal{V})$ is compact and self-adjoint, \ie $\langle v_1 , K \, v_2 \rangle_{\mathcal{V}}=\langle K \, v_1 ,  v_2 \rangle_{\mathcal{V}}$. Since  $K$ is self-adjoint, the SVD guarantees that $K$ can be decomposed as  $K=K^{\frac{1}{2}}\, K^{\frac{1}{2}} $ where $ K^{\frac{1}{2}} \in  \mathcal{B}(\mathcal{V},\mathcal{V})$, and  that  $\langle v_1 , v_2 \rangle_{\mathcal{V}_K}=\langle K^{\frac{1}{2}} \, v_1 , K^{\frac{1}{2}} \, v_2 \rangle_{\mathcal{V}}$.    The modified   Hilbert-Schmidt norm is then defined   for   $M \in \mathcal{B}(\mathcal{U},\mathcal{V})$ as 
$$
\| M \|_{\mathcal{HS},K} =\left( \sum_{i=1}^{\textrm{dim}(\mathcal{U})}\|  K^{\frac{1}{2}} \, M e^\mathcal{U}_i \|^2_\mathcal{V} \right)^{\frac{1}{2}},
$$which can be rewritten as   
$$
\| M \|_{\mathcal{HS},K} =\left( \sum_{i=1}^{m}  \sum_{j=1}^{\textrm{dim}(\mathcal{V})} \sigma_j^K (\sigma_i^M)^2 | \langle \varphi_i^M,\varphi_i^K \rangle_{\mathcal{V}}|^2   \right)^{\frac{1}{2}} .
$$
An optimal solution of problem \eqref{eq:prob1} in the sense of the norm $\| \cdot \|_{\mathcal{HS},K}$ is  then
$$M_k^\star=(K^{\frac{1}{2}})^\dagger \, P'_k \,  K^{\frac{1}{2}} \, Y \,  X^\dagger,$$
 where $P'_k=\sum_{i=1}^k \varphi_i^{Z'} \langle \varphi_i^{Z'}  , \cdot \rangle_\mathcal{V}$   with  $Z'=K^{\frac{1}{2}} \, Y \, X^\dagger  \, X$.  \\

\end{remark}
 
 \section{DMD Particularizations}
 

 \subsection{Unconstrained DMD}
If $X$ is full rank, or equivalently $\textrm{rank}(X)=m$, then $Z=Y$ and  the optimal approximation error simplifies to $$ \|Y -M_k^\star \, X \|_{\mathcal{HS}}^2=  \sum_{i=k+1}^{m} (\sigma_i^Z)^{2}   .$$ 
If $\dim(\mathcal{U})<\infty$ and $\dim(\mathcal{V})<\infty$, we recover the standard result for the unconstrained DMD problem \cite{Tu2014391}.\\

 \subsection{Low-rank DMD}

In the case $\dim(\mathcal{U})<\infty$ and $\dim(\mathcal{V})<\infty$, we recover the optimal result proposed in \cite[Theorem 4.1]{HeasHerzet17}  for low-rank DMD (or extended DMD in finite dimension). Sub-optimal solutions to this problem have been proposed in~\cite{Chen12,Jovanovic12,williams2014kernel,williams2015data} \\

 \subsection{ Kernel-Based DMD}
 
In the case $\dim(\mathcal{V})=\infty$, the result characterizes the solution of  low-rank approximation in reproducing kernel Hilbert spaces, on which  kernel-based DMD relies. Theorem~\ref{Theorem:1} justifies in this case the solution computed by the optimal kernel-based  DMD  algorithm proposed in~\cite{HeasHerzet18Kernel}. We note that  the proposed solution has been already given in \cite{williams2014kernel}  for the  infinite dimensional setting, but in the case where $k>m$. Nevertheless, the solution  provided  by the authors is sub-optimal in the general case.  \\

\subsection{Continuous DMD}\label{remark:4}
In the case $\dim(\mathcal{U})=\infty$,  the result characterizes the solution   of a continuous version of the DMD problem, where the number of snapshots are  infinite. The problem is the DMD counterpart to  the continuous POD problem presented in \cite[Theorem 6.2]{quarteroni2015reduced}. 
Here,  problem \eqref{eq:prob1} is defined as follows. $X,Y \in \mathcal{B}( \mathcal{L}^2(\mathcal{U}, \mu ) , \mathbb{R}^n)$ in \eqref{eq:prob1} are compact Hilbert-Schmidt operators, defined by their kernels $$X:  g \to  \int_\mathcal{U} k_X(u)g(u) d\mu(u) \quad  \textrm{and}\quad  Y :  g \to \int_\mathcal{U} k_Y(u)g(u) d\mu(u),
$$  where  $k_X, k_Y\in \mathcal{B}( \mathcal{L}^2(\mathcal{U}, \mu ) , \mathbb{R}^n)$ are the Hilbert-Schmidt kernels with $\mathcal{U}$ supplied by the measure $\mu$ and  $\mathcal{V}=\mathbb{R}^n$ so that $ \mathcal{B}_k(\mathcal{V},\mathcal{V})=  \{  M \in \mathbb{R}^{n \times n}: \textrm{rank}(M)\le k \} $.  The solution and the optimal error  are then characterized by Theorem~\ref{Theorem:1}.   \\

\section{Proof of Theorem~\ref{Theorem:1}}\label{ProofTH}
We will use the following extra notations in the proof. 
We define $\tilde Y= Y \,    V_X^{\rank(X)}\quad\textrm{and}\quad \tilde X= X   \, V_X^{\rank(X)}$,  where for any $k>0$, 
$V_X^k  =\sum_{i=1}^{k}   \psi_i^X \langle e_i^\mathcal{U}, \cdot \rangle_{\mathcal{U}}$. We thus have $ \tilde X= \sum_{i=1}^{\rank(X)} \sigma_i^X \varphi^X_i  \langle e_i^\mathcal{U}, \cdot \rangle_{\mathcal{U}} .$
 Finally, let 
 $
 U_{\tilde Y}^*=\sum_{i=1}^{\dim(\mathcal{V})}   e_i^{\mathcal{V}} \langle \varphi_i^{\tilde Y}, \cdot \rangle_{\mathcal{V}}$   and $ V_{\tilde Y}=\sum_{i=1}^{\dim(\mathcal{U})}   \psi_i^{\tilde Y} \langle e_i^\mathcal{U}, \cdot \rangle_{\mathcal{U}}
 $\\

\subsection{Closed-Form Solution $M_k^\star$}

 We begin by proving that  problem  \eqref{eq:prob1} admits the solution $M^\star_k$.\\

 First, we  remark that $\tilde X$ is full-rank ($\rank(\tilde X)=r$)  so that $\tilde X^\dagger \, \tilde X=I_r$,  with $r=\rank(X)$. 
Therefore, using the Pythagore Theorem, we have
\begin{align*}
\min_{M \in  \mathcal{B}_k(\mathcal{V},\mathcal{V}) } \|Y- M \, X \|_{\mathcal{HS}}^2= 
\min_{M \in  \mathcal{B}_k(\mathcal{V},\mathcal{V}) } \{& \| (Y- M \, X) \, X^\dagger \, X\|_{\mathcal{HS}}^2\\
&+  \| (Y- M \, X) \, (I-X^\dagger \, X)\|_{\mathcal{HS}}^2\}.
\end{align*}
Since we have 
$$
X \, X^\dagger \, X=X\quad \textrm{thus}\quad  \| M \, X \, (I-X^\dagger \, X)\|_{\mathcal{HS}}^2\}=0 ,
$$
we obtain
\begin{align}\label{eq:2termserror}
\min_{M \in  \mathcal{B}_k(\mathcal{V},\mathcal{V}) }& \|Y- M \, X \|_{\mathcal{HS}}^2= 
\min_{M \in  \mathcal{B}_k(\mathcal{V},\mathcal{V}) } \{ \| (Y- M \, X) \, X^\dagger \, X\|_{\mathcal{HS}}^2\nonumber \\
&\quad \quad \quad\quad \quad \quad\quad \quad \quad\quad \quad \quad\quad \quad \quad+  \| Y\, (I-X^\dagger \, X)\|_{\mathcal{HS}}^2\}\nonumber\\
= &\min_{M \in  \mathcal{B}_k(\mathcal{V},\mathcal{V}) } \{ \|  Y\, X^\dagger \, X- M \,  X\|_{\mathcal{HS}}^2+  \| Y\,  (I-X^\dagger \, X)\|_{\mathcal{HS}}^2\}\nonumber \\
= &\min_{M \in  \mathcal{B}_k(\mathcal{V},\mathcal{V}) } \{ \| \tilde Y- M \,  \tilde X\|_{\mathcal{HS}}^2+  \| Y\,  (I-X^\dagger \, X)\|_{\mathcal{HS}}^2\},
\end{align}
where the last equality follows from the invariance of the  Hilbert-Schmidt norm to unitary transforms and the fact that $ X^\dagger \, X \, V_X^r=V_X^r$.

Second, from the Sylvester inequality, we get
\begin{align*}
\min_{\tilde M \in  \mathcal{B}_k(\mathcal{U},\mathcal{V}) } \|\tilde Y- \tilde M \|_{\mathcal{HS}}^2 \le \min_{M \in  \mathcal{B}_k(\mathcal{V},\mathcal{V}) } \|\tilde Y- M \,  \tilde X \|_{\mathcal{HS}}^2,
\end{align*}
and by  invariance of the Hilbert-Schmidt norm  to unitary transforms, we obtain
\begin{align*}
\underbrace{\min_{\Lambda \in  \mathcal{B}_k(\mathcal{U},\mathcal{V}) } \|\Lambda_{\tilde Y} - \Lambda \|_{\mathcal{HS}}^2}_{(P_1)} \le \underbrace{\min_{M \in  \mathcal{B}_k(\mathcal{V},\mathcal{V}) } \|\Lambda_{\tilde Y} -  U_{\tilde Y}^* \, M \,  \tilde X \, V_{\tilde Y} \|_{\mathcal{HS}}^2}_{(P_2)},
\end{align*}
where  $\Lambda_{\tilde Y}=\sum_{i=1}^{m}    \sigma_i^{\tilde Y}  e_i^\mathcal{V} \langle e_i^\mathcal{U}, \cdot \rangle_{\mathcal{U}}$. 

Third, from the Schmidt-Eckhart-Young-Mirsky theorem \cite{schmidt1907theorie}, problem ($P_1$) admits the solution 
\begin{align*}
\Lambda_k^\star= \sum_{i=1}^{k}   \sigma_i^{\tilde Y} e_i^\mathcal{V} \langle e_i^\mathcal{U}, \cdot \rangle_{\mathcal{U}}
\end{align*} 

Fourth, we  remark  that
 $V_X^r  \, \tilde X ^\dagger=X^\dagger$
 and that 
 the truncation to $k$ terms of the SVD of $X ^\dagger \,  X$ corresponds to the operator $V^k_X$ yielding $ P_k=Y \, V^k_X \, (V^k_X)^* \, Y^* $.
 Therefore, $M_k^\star=Y \, V^k_X \, (V^k_X)^* \, Y^* \,  \tilde Y \, \tilde X^\dagger$ and we verify that $$U_{\tilde Y}^*\, M_k^\star\,  \tilde X\,   V_{\tilde Y}=\Lambda_k^\star.$$
We deduce that the minimum of the objective function of ($P_2$)   reaches the minimum of the  objective function of ($P_1$) at  $M=M_k^\star$, \ie
$$
\|\Lambda_{\tilde Y} -  U_{\tilde Y}^* \, M_k^\star \,  \tilde X \, V_{\tilde Y} \|_{\mathcal{HS}}^2=\|\Lambda_{\tilde Y} - \Lambda_k^\star \|_{\mathcal{HS}}^2,
$$

Finally, since the objective function of ($P_2$)  reaches at $M_k^\star$ its lower bound, $M_k^\star$ is a minimiser of ($P_2$). We then deduce from \eqref{eq:2termserror}, that $M_k^\star$ is also a minimiser of problem \eqref{eq:prob1}. $\square$

\subsection{Characterization of the Optimal Error Norm}
It remains to characterize the  error norm \eqref{eq:optError}. 
On the one hand, we have
\begin{align*}
Y\,  (I-X^\dagger \, X)= \sum_{i=1}^{m}\sum_{j=r+1}^{m}\sigma_i^Y \varphi_i^Y \langle \psi_j^X, \cdot \rangle_{\mathcal{U}} \langle  \psi_i^Y,  \psi_j^X \rangle_{\mathcal{U}}.
\end{align*}
Since $\{\psi_\ell^X\}_{\ell>0}$ is a ONB of $\mathcal{U}$, we can expand the norm and obtain
\begin{align*}
 \| Y\,  (I-X^\dagger \, X)\|_{\mathcal{HS}}^2 &=\sum_{\ell=1}^{\dim(\mathcal{U})}  \| Y \, (I-X^\dagger \, X)  \psi_\ell^X \|^2_{\mathcal{V}} ,  \\
 &=\sum_{\ell=1}^{\dim(\mathcal{U})}  \| \sum_{i=1}^{m}\sum_{j=r+1}^{m}\sigma_i^Y \varphi_i^Y \langle \psi_j^X, \psi_\ell^X \rangle_{\mathcal{U}} \langle  \psi_i^Y,  \psi_j^X \rangle_{\mathcal{U}} \|^2_{\mathcal{V}} ,\\
  &=\sum_{\ell=r+1}^{m}  \| \sum_{i=1}^{m}\sigma_i^Y \varphi_i^Y  \langle  \psi_i^Y,  \psi_\ell^X \rangle_{\mathcal{U}} \|^2_{\mathcal{V}} ,\\
   &= \sum_{\ell=r+1}^{m}   \| \sum_{i=1}^{m}\sigma_i^Y \varphi_i^Y  \langle  \psi_i^Y,  \psi_\ell^X \rangle_{\mathcal{U}} \|^2_{\mathcal{V}},\\
   &= \sum_{\ell=r+1}^{m}  \sum_{i=1}^{m} \| \sigma_i^Y \varphi_i^Y  \langle  \psi_i^Y,  \psi_\ell^X \rangle_{\mathcal{U}} \|^2_{\mathcal{V}},\\
   &= \sum_{\ell=r+1}^{m}  \sum_{i=1}^{m} ( \sigma_i^Y  \langle  \psi_i^Y,  \psi_\ell^X \rangle_{\mathcal{U}} )^2 ,
\end{align*}
where, in order to obtain the two last equalities, we have exploited the fact that  $\{ \varphi_i^Y \}_{i>0}$ is an ONB of $\mathcal{V}$.
On the other hand, we have 
\begin{align*}
\|\Lambda_{\tilde Y} - \Lambda_k^\star \|_{\mathcal{HS}}^2&=\sum_{j=1}^{\dim(\mathcal{U})} \| \sum_{i=1}^{m}   \sigma_i^{\tilde Y} e_i^\mathcal{V} \langle e_i^\mathcal{U}, e_j^\mathcal{U} \rangle_{\mathcal{U}}   - \sum_{i=1}^{k}   \sigma_i^{\tilde Y} e_i^\mathcal{V} \langle e_i^\mathcal{U}, e_j^\mathcal{U} \rangle_{\mathcal{U}}\|^2_{\mathcal{V}},\\
&= \sum_{j=1}^{\dim(\mathcal{U})} \| \sum_{i=k+1}^{m}   \sigma_i^{\tilde Y} e_i^\mathcal{V} \langle e_i^\mathcal{U}, e_j^\mathcal{U} \rangle_{\mathcal{U}}  \|^2_{\mathcal{V}},\\
&= \sum_{i=k+1}^{m}   \|    \sigma_i^{\tilde Y} e_i^\mathcal{V}   \|^2_{\mathcal{V}},\\
&= \sum_{i=k+1}^{m}   (\sigma_i^{\tilde Y})^2.
\end{align*}

Finally, from \eqref{eq:2termserror} and the two above expressions, we conclude
\begin{align*}
 \|Y- M_k^\star \, X \|_{\mathcal{HS}}^2&= \| \tilde Y-  M_k^\star \,  \tilde X\|_{\mathcal{HS}}^2+  \| Y\,  (I-X^\dagger \, X)\|_{\mathcal{HS}}^2,\\
&=\|\Lambda_{\tilde Y} - \Lambda_k^\star \|_{\mathcal{HS}}^2+  \| Y\,  (I-X^\dagger \, X)\|_{\mathcal{HS}}^2,\\
&=\sum_{i=k+1}^{m}   (\sigma_i^{\tilde Y})^2+\sum_{\ell=r+1}^{m}  \sum_{i=1}^{m} ( \sigma_i^Y  \langle  \psi_i^Y,  \psi_\ell^X \rangle_{\mathcal{U}} )^2 .\quad \square
\end{align*}

\section{Conclusion}

We have  shown that there exists a closed-form optimal solution to the non-convex problem related to low-rank approximation of linear bounded operators in the sense of the Hilbert-Schmidt norm. 
 This result generalizes to low-rank operator in Hilbert spaces  solutions obtained in the context of low-rank matrix approximation. As in the latter finite-dimensional case, the proposed closed-form  solution takes the form of the orthogonal projection of the solution of the unconstrained problem onto a specific low-dimensional  subspace. However, the proof is substantially different. It relies on the well-known  Schmidt-Eckhart-Young-Mirsky theorem.  
The proposed theorem is  discussed and applied to various contexts, including  low-rank approximation for  kernel-based or continuous   DMD.

 \bibliographystyle{spmpsci}     
\bibliography{./bibtex}

\end{document}